\documentclass[final]{cvpr}

\usepackage{times}
\usepackage{epsfig}
\usepackage{graphicx}
\usepackage{amsmath}
\usepackage{amssymb}
\usepackage{float}
\usepackage{caption}

\usepackage[pagebackref=true,breaklinks=true,colorlinks,bookmarks=false]{hyperref}

\def\cvprPaperID{1084} 
\def\confYear{CVPR 2021}

\DeclareMathOperator*{\argmin}{\arg\min}

\def\L{{\cal L}}

\def\dE{{\mathbb{E}}}

\def \TODO [#1]{\textcolor{red}{TODO: #1}}
\def \NOTE [#1]{\textcolor{blue}{(\textit{#1})}}
\def \comment [#1]{}
\def \etal {\emph{et al.~}}

\begin{document}
	
\title{Intrinsic Temporal Regularization for High-resolution Human Video Synthesis}

\author{Lingbo Yang$^1,2$, Zhanning Gao$^2$, Peiran Ren$^2$, Siwei Ma$^1$, Wen Gao$^1$\\
	1. Institute of Digital Media, School of EECS, Peking University\\
	2. Alibaba DAMO Academy\\
	{\tt\small \{lingbo, swma, wgao\}@pku.edu.cn  \{zhanning.gzn, peiran.rpr\}@alibaba-inc.com}
}

\twocolumn[{
	\maketitle
	\begin{center}
		\centering
		\includegraphics[width=.95\textwidth]{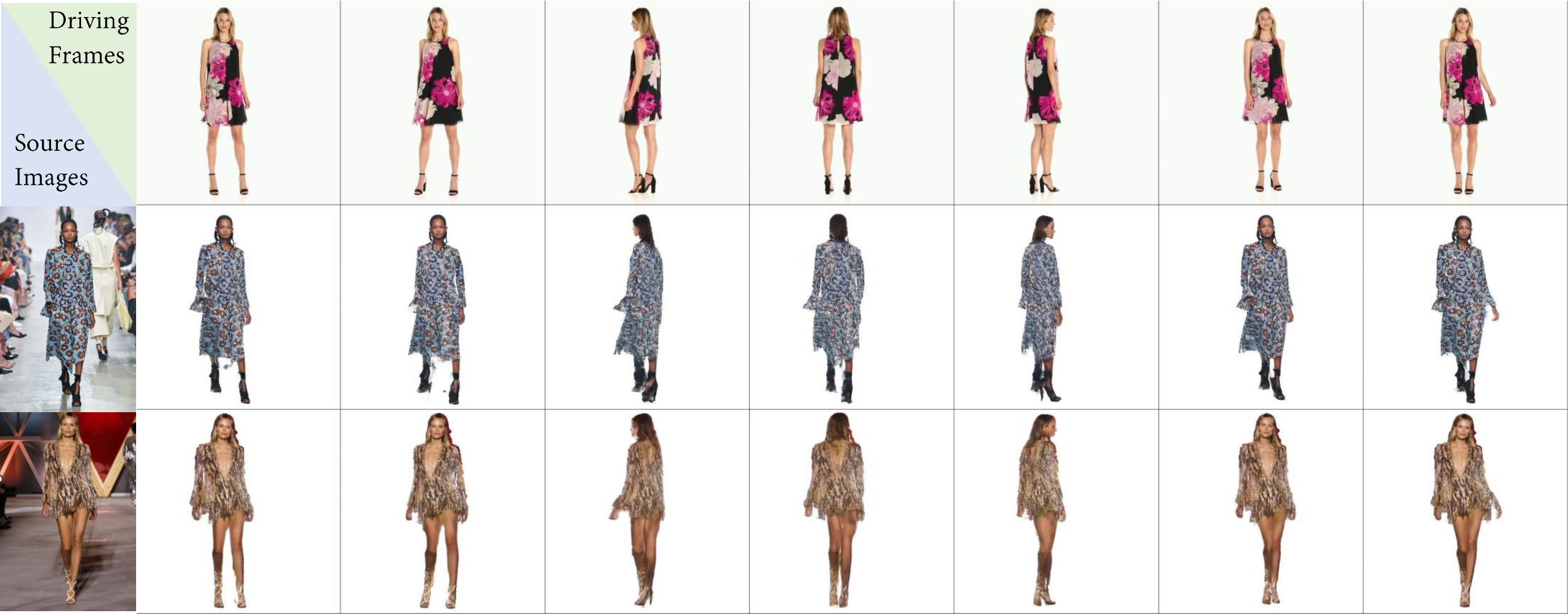}
		\captionof{figure}{Performances of our method over real-world runway shows. \emph{Best to zoom in for details.}}
		\label{fig:teaser}
	\end{center}%
}]

\begin{abstract}
	Temporal consistency is crucial for extending image processing pipelines to the video domain, which is often enforced with flow-based warping error over adjacent frames. Yet for human video synthesis, such scheme is less reliable due to the misalignment between source and target video as well as the difficulty in accurate flow estimation. In this paper, we propose an effective intrinsic temporal regularization scheme to mitigate these issues, where an intrinsic confidence map is estimated via the frame generator to regulate motion estimation via temporal loss modulation. This creates a shortcut for back-propagating temporal loss gradients directly to the front-end motion estimator, thus improving training stability and temporal coherence in output videos. We apply our intrinsic temporal regulation to single-image generator, leading to a powerful ``INTERnet'' capable of generating $512\times512$ resolution human action videos with temporal-coherent, realistic visual details. Extensive experiments demonstrate the superiority of proposed INTERnet over several competitive baselines.
\end{abstract}

\section{Introduction}

Human video synthesis is an emerging topic with huge potential in human-centered applications, such as fashion design~\cite{FWGAN}, media production~\cite{EverybodyDanceNow}, virtual reality, and so on. For these applications, the goal is to synthesis videos of a specific person from a given source image by incorporating motion guidance from another driving video, where temporal consistency plays a crucial role in promoting user experience.
Despite the recent progress in human image generation~\cite{DeformableGAN}\cite{GFLA}\cite{MyICME2020}\cite{MyTIP2020}, extending these methods to create high-resolution, temporal-coherent videos still remains a considerable challenge.

More specifically, existing methods for human image generation usually follow a two-stage ``warping-based'' approach: Stage I aims to estimate a dense motion field between source image and driving frame from extracted pose representations, such as sparse keypoints~\cite{DeformableGAN}, body parsing maps~\cite{SPT}\cite{softgate}, or 3D surface coordinates~\cite{DensePoseTransfer}\cite{coordinate-based}\cite{DIAF}\cite{liquidwarpinggan}. Thereafter, stage II generates target person images based on appearance features extracted from the source image, which have been warped to the target pose according to the estimated motion field. Multi-scale feature fusion~\cite{pix2pixHD} and attention mechanism~\cite{PATN}\cite{GFLA}\cite{cocosnet} are usually adopted in this stage to hallucinate visual contents in occluded regions. However, due to the non-rigid human body deformation and stochastic fabric dynamics in loose garments (like dresses), motion field estimation is often temporally unstable, which inevitably brings severe flickering between adjacent frames.

\begin{figure}
	\centering
	\includegraphics[width=1\linewidth]{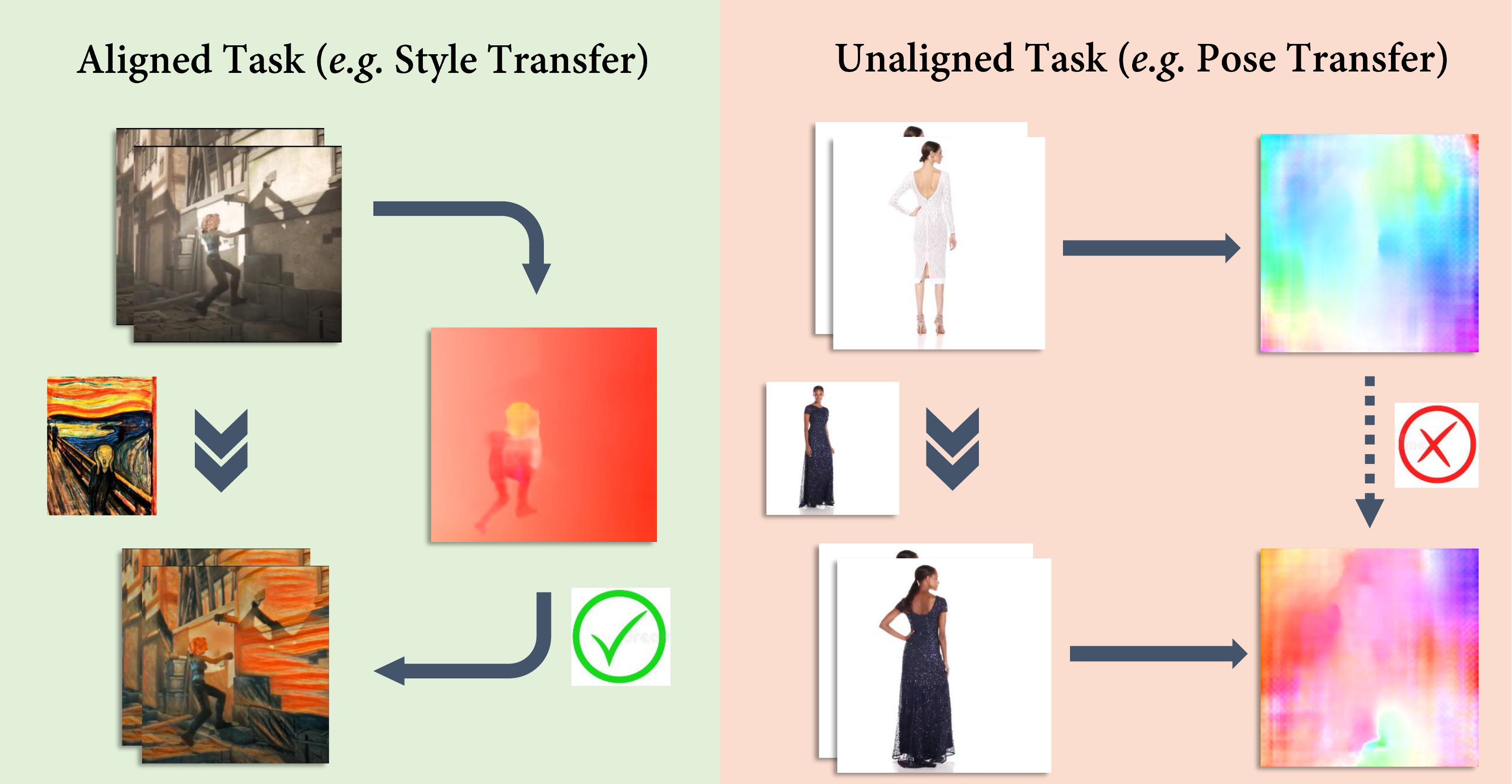}
	\caption[intrinsic reg]{\textbf{The fundamental challenge in human video synthesis.} Most temporal consistency learning methods rely on exact motion correspondence between source and target videos, which is not guaranteed for human videos.}
	\label{fig:problem}
\end{figure}

Temporal consistency learning aims to extend single-image processing methods to the video domain with extrinsic guidance extracted from previous frames, where many recent works focus on learning blind video temporal consistency~\cite{LearningBlindTemporal}~\cite{DeepVideoPrior} across different tasks, such as style transfer~\cite{CoherentVST}, enhancement~\cite{VideoEnhan} and video-to-video synthesis~\cite{vid2vid}.
Here a critical assumption is the exact structural alignment between source and target video frames, so that the required pixel-level warping flow can be directly estimated from the source video. Yet such assumption is usually violated in human videos: Fig.~\ref{fig:problem} illustrates this behavior with optical flow estimated from two different videos, where pose similarity does not necessarily imply pixel-level alignment, leading to huge flow discrepancy between different persons. This leads to a natural question: \emph{How to promote temporal consistency between generated frames without accurate extrinsic motion guidance?}


In this paper, we propose an intrinsic temporal regularization scheme to address the above challenge in high-resolution human video synthesis. Instead of regulating with extrinsic flow, we propose to modulate the traditional flow-based warping loss with an intrinsic confidence map predicted from the motion field estimator. Our design heuristics are twofold: Theoretically, we establish the connection between stability in motion estimation and temporal coherence in generated video frames; and technically, we create an effective shortcut for temporal loss gradients to be directly back-propagated into Stage-I motion estimator (instead of filtering through Stage-II frame generator), which improves the training stability and robustness of motion estimation between source image and driving frames. Our intrinsic regularization scheme can be easily integrated into existing human image generation frameworks with little computational overhead, and is capable of generating realistic $512\times512$ video clips based on real-world person images, for instance, snapshots of runway models (Fig.~\ref{fig:teaser}). Overall, our contributions are threefold:

\begin{itemize}
	\item We propose an effective intrinsic regularization scheme to address the key challenge in temporal consistency learning for human video synthesis, where ground truth motion between target frames cannot be accurately estimated from either the source video or reference frames.
	\item We implement an intrinsic temporal regularization network (dubbed as ``INTERnet'' hereafter) to tackle high-resolution human video synthesis with little computational overhead against frame-wise generation.
	\item Extensive experiments over benchmark dataset verifies the superiority of our INTERnet against existing methods in generating realistic human action videos with more detailed textures and better temporal consistency.
\end{itemize}

\section{Related Works}

\textbf{Single-image Human Pose Transfer} aims to generate new person images based on a given source image and the target pose, where existing works roughly boil down to two categories. Transform-based methods aim to infer a pixel-wise feature warping based on given pose representations, including local affine transform~\cite{DeformableGAN}\cite{first-order-motion}, thin-plate splines~\cite{softgate}\cite{VITON}\cite{CPVTON}, optical flow~\cite{vid2vid}\cite{unsupervisedPF}, and 3D surface models~\cite{DIAF}\cite{DensePoseTransfer}\cite{coordinate-based};
Attention-based methods aim to capture feature correlations between source and target images in varying scopes, including local~\cite{GFLA}, global~\cite{guidedpix2pix}\cite{cocosnet}, and progressive attention~\cite{PATN}.  Moreover, Yang~\etal~\cite{MyICME2020}\cite{MyTIP2020} introduce a residual texture enhancement module to refine clothing and facial details, and Men~\etal~\cite{ADGAN} further decompose the attributes across different semantic regions for better user control.
Despite the achieved progress, there still remains considerable challenges in generating high-resolution person images and extending current methods to the video domain.

\textbf{Temporal Consistency Regularization} is usually enforced at either feature level or loss level. For feature regulation, extrinsic motion cues are often provided to warp features from previous frames into the current generative cycle, which often involve recurrent networks for temporal relation modeling: Tulyakov~\etal propose MoCoGAN~\cite{MoCoGAN} to predict video frames based on a constant content vector and a stochastic motion vector sequence, and Chen~\etal\cite{CoherentVST} pose coherent video style transfer by fusing feature maps warped from the previous generated frame, where the warping flow is estimated from the source video. Yet feature regulation is often infeasible due to the misalignment between source and target videos, as shown in Fig.~\ref{fig:problem}.
Another approach is by regulating the loss between adjacent frames, such as an adversarial loss~\cite{EverybodyDanceNow}\cite{DeepFashion}\cite{MyICME2019} or backwarp loss between spatially-aligned frames from pre-computed optical flow~\cite{vid2vid}\cite{fewshotvid2vid}. However, optical flow estimation for human videos with large displacements and textureless background is often inaccurate~\cite{BadFlow}, which could end up affecting the temporal regularization learning. In this paper, we propose to modulate the flow-based backwarp loss via an intrinsic confidence map associated with motion field estimation, leading to more stable feature transformation and better temporal consistency in resulted videos.

\section{Intrinsic Temporal Regularization}
\label{sec:itr}

In this section, we first introduce the general formulation of human video synthesis, and analyze two existing temporal regularization schemes. This motivates the design of our intrinsic regularization scheme, which involves modulating flow-based warping error with an intrinsic confidence map associated with estimated motion field. Finally, we apply our intrinsic regularization to existing human image generation networks and obtain ``INTERnet''.

\subsection{Formulation and Analysis}

Given a source image $S$ and a driving video $\mathbf{D} = [D_1;D_2;\dots;D_n]$, human video synthesis aims to produce a new video clip $\mathbf{O} = [O_1;O_2;\dots;O_n]$ that matches the person's appearance in $S$ and the body movement in $\mathbf{D}$. Generally, it involves training a human image generator $G$ to process each driving frame sequentially, so that $O_t = G(S;D_t)$. Given the ground truth reference video $\mathbf{I} = [I_1;I_2;\dots;I_n]$ for $S$, we can formulate the optimization of each individual frame as follows:


\begin{equation}
O_t =\argmin_{O}(\L_{frame}(O;I_t) + \lambda_{reg}\L_{reg}(O;I_t,I_{t-1}))
\label{eqn:overall}
\end{equation}

\begin{figure*}
	\centering
	\includegraphics[width=1\linewidth]{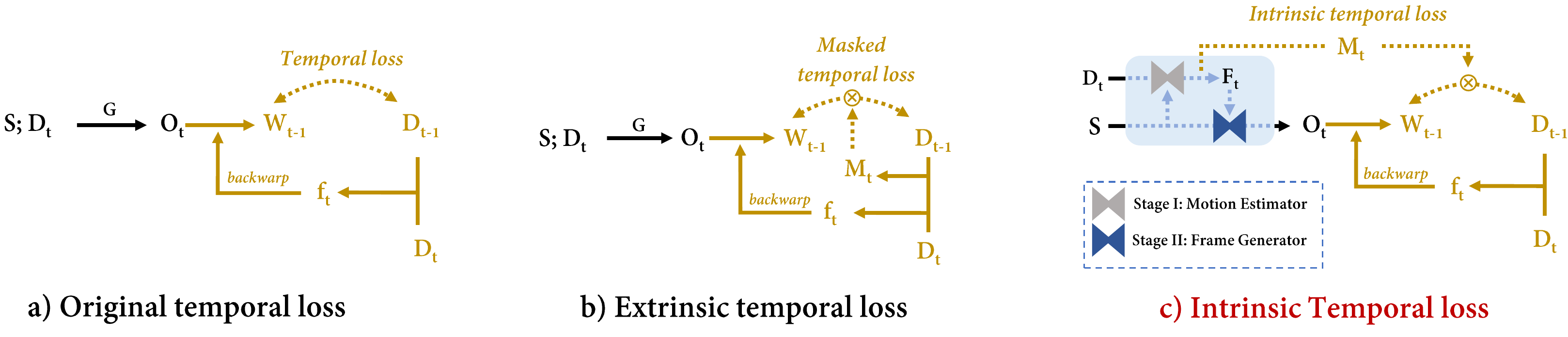}
	\caption[intrinsic reg]{Illustration of different temporal regularization schemes. The loss computation module is marked in gold color.}
	\label{fig:intrinsicreg}
\end{figure*}

Here $\L_{reg}$ denotes the regularization term for enforcing temporal consistency between current frame and the previous frame. Early works in human video generation attempt to capture the temporal relationship implicitly via adversarial loss~\cite{vid2vid}\cite{EverybodyDanceNow}, which are less effective in modeling pixel-level consistency. Instead, the most widely adopted regulation scheme is the following \emph{flow warp loss} (Fig.~\ref{fig:intrinsicreg}a):

\begin{equation}
\L_{reg} = \|f_t(O_t) - I_{t-1}\|
\label{eqn:flow}
\end{equation}

\begin{figure}
	\centering
	\includegraphics[width=1.0\linewidth]{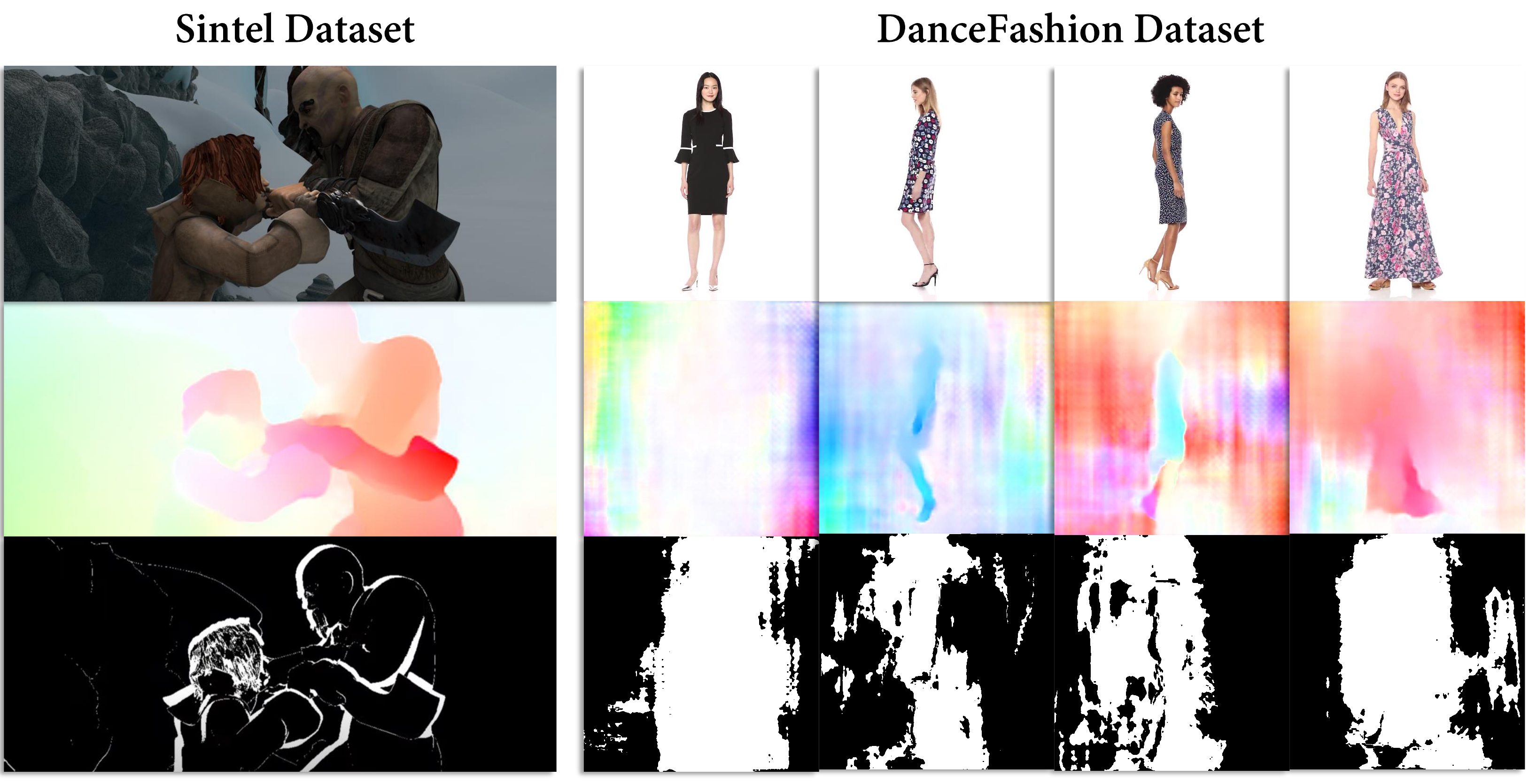}
	\caption[badflow]{Challenges in optical flow estimation and occlusion detection for human video synthesis. While our flow estimator behaves normal on common benchmark datasets, it tends to incur severe background noises for human videos, which also affects occlusion mask detection. }
	\label{fig:badflow}
\end{figure}

Here $f_t$ is the ground truth optical flow between $I_t$ and $I_{t-1}$, which is applied to align current generated frame $O_t$ with previous ground truth frame $I_{t-1}$ via back-warping. Needless to say, accurate optical flow is critical for reliable temporal consistency learning.
Yet flow estimation can be quite challenging for human videos with fast movement, frequent occlusion/motion blur and textureless background~\cite{opticalflowsurvey}, making the corresponding warping loss less reliable for our task at hand. Fig.~\ref{fig:badflow} compares the flow field estimated upon the benchmark Sintel dataset~\cite{sintel} and several human video clips, where the latter case involves severe background noises.
Therefore, an extrinsic confidence map 
is usually determined from adjacent frames to compensate for such inaccuracy (Fig.~\ref{fig:intrinsicreg}b):

\begin{equation}
\L_{reg}^{ext} = \|(f_t(O_t) - I_{t-1}) \cdot M_t\|
\label{eqn:ext_flow}
\end{equation}

The confidence map $M_t$ indicates the reliability of motion estimation at each pixel of the current frame. So far, the most widely-used confidence criteria is defined in~\cite{flow_refine_criteria} that aims to exclude two types of pixels: 1) The occluded ones that are inconsistent between forward and backward flow, and 2) the ones at motion boundaries with large spatial gradients. Yet due to the inaccurate flow estimation, the acquired map $M_t$ still contains shattered object boundaries and background residue, as shown in Fig.~\ref{fig:badflow} bottom row.

\subsection{An Intrinsic Approach}
To address the above issue in flow-based regularization, we propose an intrinsic approach to estimate the confidence map not from generated frames, but from the front-end motion field estimator.
As shown in Fig.~\ref{fig:intrinsicreg}, existing human image generation methods typically involve an explicit motion field estimation $F_t = F(S; D_t)$ from the inputs that are utilized to warp source image features under the target pose. This motivates us to also estimate a confidence map $M_t$ from the motion estimator, which can be readily implemented by attaching an additional layer to the last feature map of $F_t$. Thus the proposed \emph{intrinsic temporal regularization loss} is formulated as:

\begin{equation}
\L_{reg}^{int} = \|(f_t(O_t) - I_{t-1}) \cdot M_t(F_t)\|
\label{eqn:int_flow}
\end{equation}

Compared to Eqn.~\eqref{eqn:flow} and~\eqref{eqn:ext_flow}, the proposed intrinsic loss establishes the connection between motion field estimation and optical flow compensation.
More precisely, let us denote the human body mask in source and driving frame as $\Omega_s$ and $\Omega_{D_t}$, then any estimated motion field $F: \Omega_s \rightarrow \Omega_{D_t}$ naturally derives a set $\Omega_{F} = \Omega_s\backslash F^{-1}(\Omega_{D_t})$ of occluded pixels that are not directly warped from the source image $S$, but hallucinated from stage II generator. As such, the generated contents in $\Omega_{F}$ share less correlation with the input source image (and hence any other reference frame), which is more likely to cause inaccurate optical flow estimation. Therefore, it would be reasonable to mask out this region when calculating flow-based temporal regularization loss to mitigate the impact of hallucinated contents. The rationality of our analysis will be empirically verified in ablation study, Sec.~\ref{sec:ablation}. 

\subsection{Building the ``INTERnet''}
We substantiate the conceptualized model in Fig.~\ref{fig:intrinsicreg}c) with a complete end-to-end framework dubbed ``INTERnet''. The overall architecture is shown in Fig.~\ref{fig:framework}. Below we detail the implementation of each module:

\begin{figure*}[t]
	\centering
	\includegraphics[width=1\linewidth]{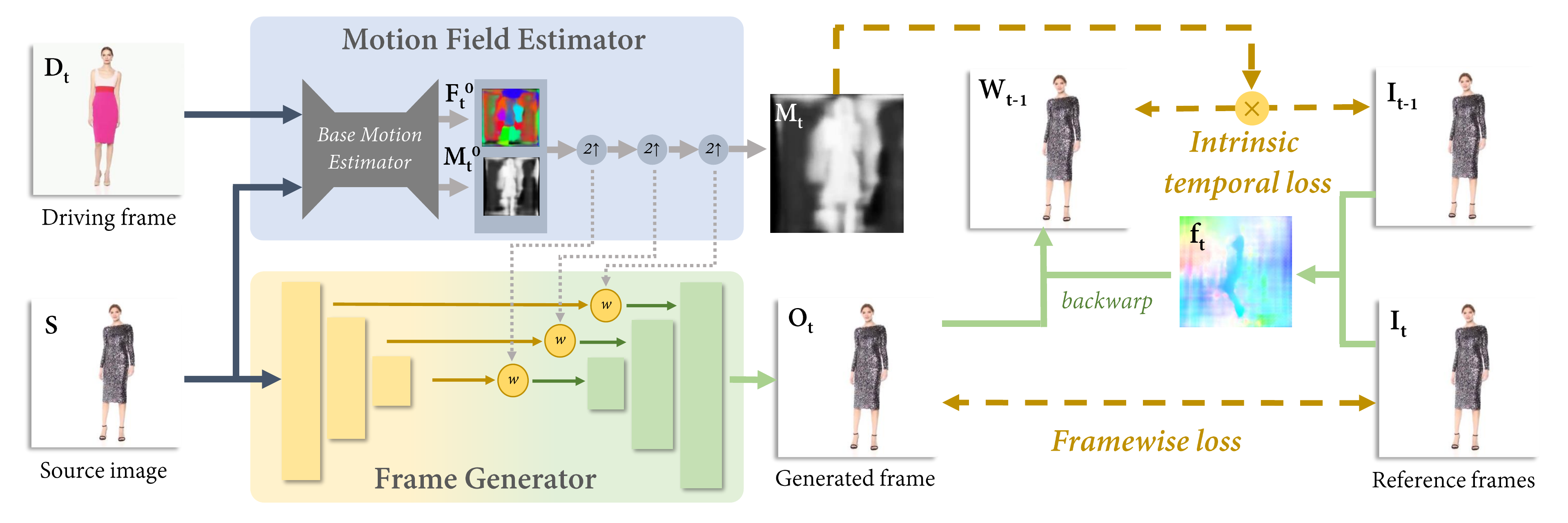}
	\caption[framework]{\textbf{Overview of the proposed framework.} The motion field of image features from the source image to the driving frame are first estimated at the smallest scale, then progressively upsampled. The acquired multi-scale motion fields are then utilized to warp corresponding features of the source image for progressive frame generation. In addition, an intrinsic confidence map is to regularize the backwarp loss between adjacent frames, which compensates for the inaccuracy in optical flow and stabilizes the training of base motion estimator.}
	\label{fig:framework}
\end{figure*}

\textbf{Stage I: Motion Field Estimation. } We adopt the first-order-motion model~\cite{first-order-motion} to decompose the global motion field $F_t$ into a set of local affine transforms $T_i$:

\begin{align}
F_t &= \sum_{i=1}^{n} \omega_i T_i \\
s.t.\quad T_i(K^s_i) &= K^d_i, \sum_{i=1}^{n} \omega_i = 1 \nonumber
\end{align}

Here $K^s_i$ and $K^d_i$ denotes motion anchors from the source and driving frame, respectively, and $\{\omega_i\}$ are soft-thresholded composition weights to ensure smoothness of the combined motion field. During training, the motion anchors $K_i$, local transformation $T_i$, and the composition weights $\omega_i$ are jointly predicted via an end-to-end motion estimator. To allow efficient computation of multiscale motion, we first predict the base motion field $F_t^0$ and the corresponding confidence map $M_t^0$ at the smallest scale, then progressively scale them with 2x bilinear upsampling to get two pyramids $\{F_t^i\}$ and $\{M_t^i\}$ to allow multiscale feature transfer and fusion for high-resolution frame generation.

\textbf{Stage II: Frame Generation. } We adopt a multiscale residual network with deformable skip connections~\cite{DeformableGAN} to fuse warped source features at different scales progressively. Let $\phi^{in}_i$ denote the encoded source feature map at $i$-th layer, the deformed feature map is then computed by:

\begin{equation}
\phi^{out}_i = M_t^i \cdot F_t^i(\phi^{in}_i)
\label{eqn:feat_warp}
\end{equation}

Here $M_t^i$ is the upsampled version of $M_t^0$. Afterwards, $\phi^{out}_i$ is concatenated with the output of previous decoding layer, and sent into subsequent decoding layers to acquire the final output frame $O_t$.

%
%
%
%
%
%

\textbf{Learning objective.} The overall objective and the intrinsic temporal loss are defined in {eqn:overall} and~\eqref{eqn:int_flow}, and framewise loss is defined as:
\begin{equation}
\L_{frame} = \L_{adv} + \lambda_{per}\L_{per} + \lambda_{sty}\L_{sty}
\end{equation}

Here we adopt the multi-scale adversarial loss in~\cite{pix2pixHD} to approximate the distribution of real video frames:
\begin{equation}
\L_{adv} = \sum_{i=1}^{N_D}\dE[\log[1 - D_i(O_t)]] + \dE[\log[D_i(I_t)]]
\end{equation}

Also, perceptual loss and style loss~\cite{PerceptualLoss} are defined as:

\begin{equation}
\L_{per} = \frac{1}{CHW}\sum_{l}\|\phi_l(I_t) - \phi_l(\tilde{I}_{co})\|
\end{equation}
\begin{equation}
\L_{sty} = \frac{1}{CHW}\sum_l\|\mathcal{G}(\phi_l(I_t)) - \mathcal{G}(\phi_l(\tilde{I}_t))\|_F^2
\end{equation}

Here $\phi$ is a pretrained VGG-19 network with $l$ denoting layer indices, and $\mathcal{G}$ is the Gram matrix:
\begin{equation}
\mathcal{G}(F)_{ij} = \frac{1}{CHW}\sum_{h=1}^{H}\sum_{w=1}^W F_{ihw}F_{jhw}
\end{equation}

\textbf{Training and Inference.} Due to the lack of aligned human videos, we choose the source image $S$ to be the first frame $D_1$ from any video, and train the generator to reconstruct a randomly-selected frame $D_t$ from the same video, where $D_{t-1}$ is also applied for temporal loss computation. To focus on cases with large displacements, the driving frame is selected to be at least 2 seconds away from the beginning, so that $t > 60$. During inference, the output video clip is generated by conditioning on each driving frame with no additional cost for optical flow estimation or inter-frame warping.

\begin{figure*}
	\centering
	\includegraphics[width=1.0\linewidth]{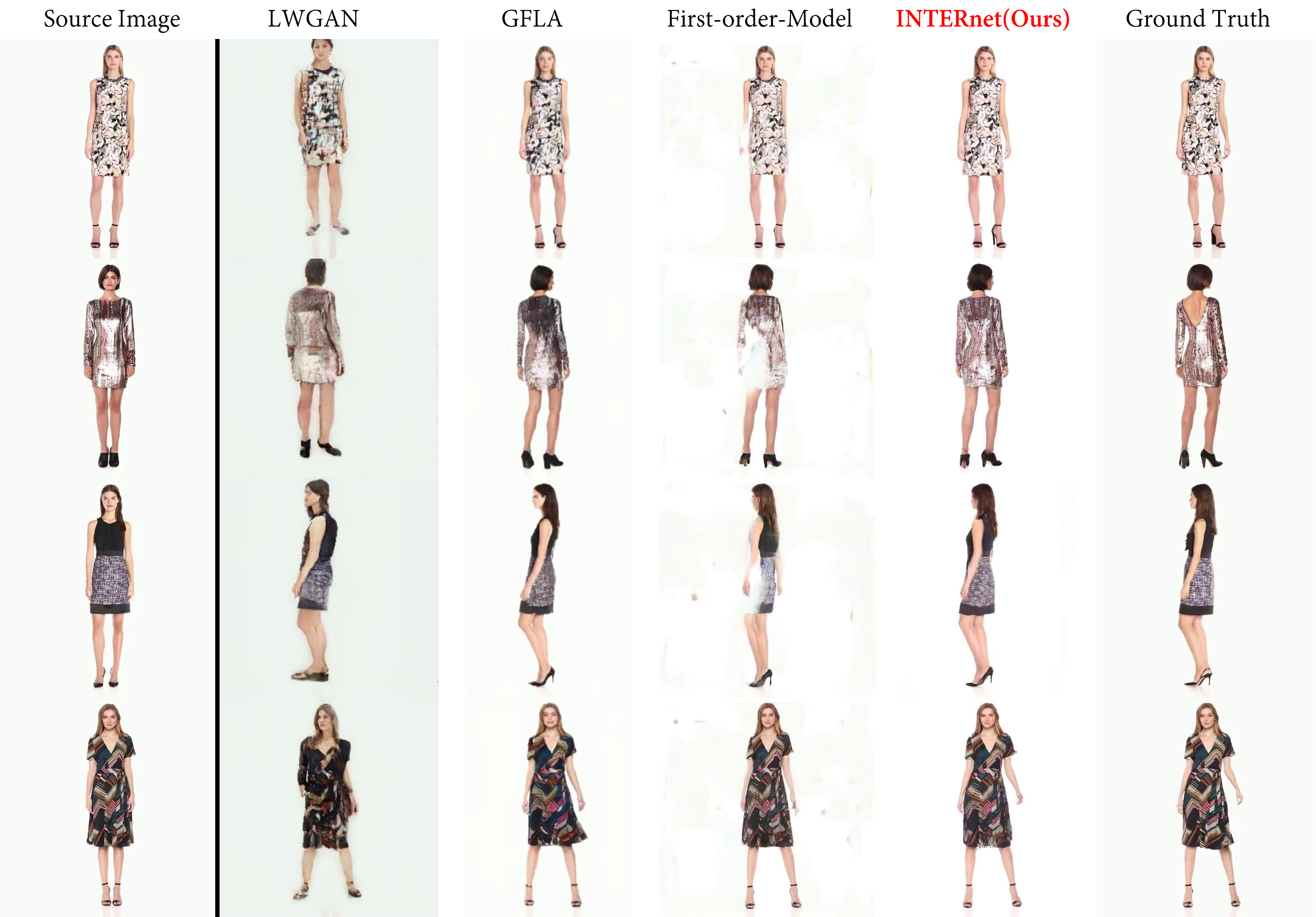}
	\caption[sota]{Qualitative reconstruction results against other baselines. Our proposed intrinsic learning scheme better preserves structural integrity and complex garment textures.}
	\label{fig:sota}
\end{figure*}

\section{Experiments}

In this section, we evaluate our proposed INTERnet against several competitive human image/video generation baselines, and perform an ablation study to verify the efficacy of our major contributions. More illustrative video generation results are provided in supplementary material.

\subsection{Experimental Settings}
\label{sec:exp_setting}

\textbf{Datasets.} We evaluate our proposed framework on the DanceFashion~\cite{dwnet} dataset containing 600 videos with 250K frames of professional models posing in diverse garments. We resize all frames to $512\times512$ and follow the default train/test splits prescribed in~\cite{first-order-motion}, with 500 videos for training and the rest 100 for testing.

\textbf{Evaluation Protocols.} We conduct both self-imitation and cross-imitation tests for each method. For self-imitation, the first frame is selected as source image to reconstruct the entire video clip, where quantitative performance is evaluated with frame-wise SSIM, LPIPS~\cite{LPIPS} and clip-wise FID~\cite{FID} metrics. For cross-imitation, we transfer the motion from driving video onto the image of another individual. Due to the lack of corresponding ground truth, we mainly resort to qualitative comparison of generated frames by conducting a user study to determine the preference between different methods.

For temporal evaluation, we design two new protocols to measure both inter-frame and long-term the consistency of generated videos. First, we employ a pretrained video frame interpolation model~\cite{SuperSloMo} to predict each frame $I_t$ from $O_{t-1}$ and $O_{t+1}$, allowing inter-frame consistency to be reflected by average PSNR. Second, we calculate the feature distance between generated and real videos via a pretrained action recognition model~\cite{I3D}, which measures the motion consistency at a semantic level. The corresponding metrics are denoted as \emph{Interp-PSNR} and \emph{Act-Recog-L1}, respectively.

\textbf{Implementation Details.} We adopt a U-net~\cite{unet} to estimate base motion field  at~$128\times128$ base resolution, with 10 motion anchors for all experiments. For frame generation, a 3-stage resnet generator~\cite{pix2pixHD} is adopted to receive warped features at~$128, 256$ and $512$ resolution. For equivalance loss, the random transformation $\mathcal{T}$ is chosen as thin-plate splines~\cite{TPS}. We train for 100 epochs using a batch of size 16 spreading across 8 Nvidia V100 cards, which takes around 24$\sim$36 hours. RAdam~\cite{radam} optimizer is adopted with initial learning rate $lr=0.0002$ and $\alpha=0.1$ decay at the end of epoch 60 and 90.

\begin{figure*}
	\centering
	\includegraphics[width=1.0\linewidth]{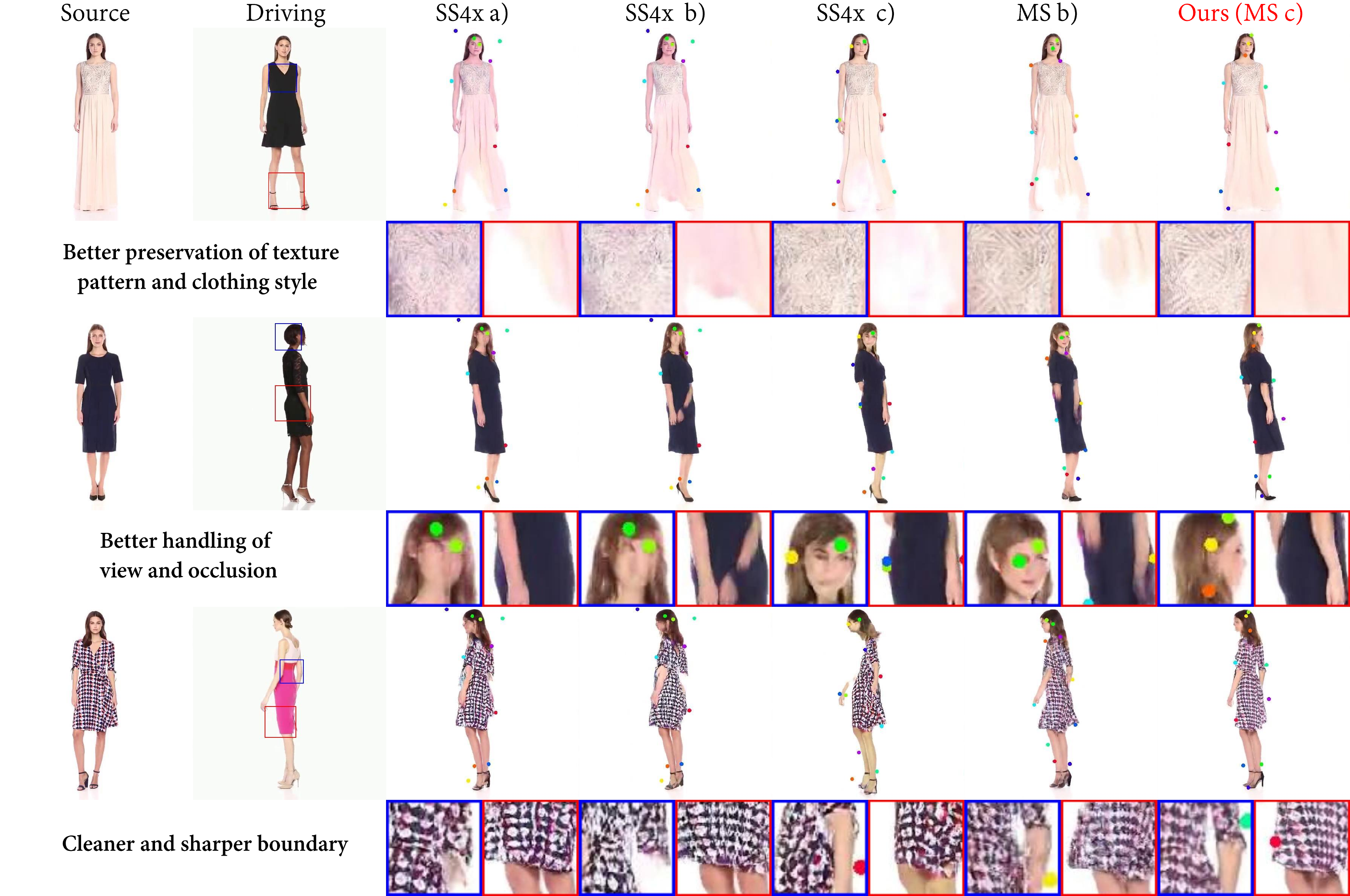}
	\caption[ablation]{Qualitative results of ablation methods. Our intrinsic temporal regularization leads to better texture details and structural integrity. Detailed regions are enlarged below each case.}
	\label{fig:ablationv1}
\end{figure*}

\subsection{Comparison with Existing Works}
\label{sec:sota}

We compare our method with three state-of-the-art baselines: Liquid warping GAN (LWGAN)~\cite{liquidwarpinggan}, global-flow-local-attention (GFLA)~\cite{GFLA}, and first-order-motion model~(FoM)~\cite{first-order-motion}.
Here LWGAN adopts a pre-built 3D human mesh template to calculate motion field, while GFLA and FoM both adopts end-to-end motion estimation with deep neural networks, and applies adversarial loss and equivariance loss for temporal regularization, respectively.
For LWGAN and GFLA, training at $512\times512$ resolution is intractable due to hardware constraints, thus we evaluate with their officially released checkpoints at $256\times256$ and rescale the output videos to the desired resolution.
Table~\ref{tab:scores} shows the quantitative scores of different methods in self-imitation task. Our INTERnet surpasses all baselines by a significant margin, demonstrating the efficacy of proposed framework. We also compare qualitative reconstruction results in Fig.\ref{fig:sota}, where LWGAN incurs severe temporal flickering across frames due to unstable 3D mesh regression, and is less capable of modeling the geometric structure of clothes. GFLA better preserves the body contour but incurs severe temporal flickering at textured regions, and FoM trained at $512\times512$ manages to preserve fine textures, yet suffer from incomplete body parts and background residue due to inaccurate motion estimation.
In contrast, our INTERnet produce much finer visual details while simultaneously reduces temporal flicking, leading to smooth and visually pleasant action video clips. \emph{Please watch our supplementary video for better comparison of temporal quality.}

\begin{table}
	\centering
	\begin{tabular}{|c|cccc|}
		\hline
		& LWGAN & GFLA & FoM & Ours\\\hline
		JND (\%) & 5.7 & 18.9 & 24.5 & \textbf{32.1}\\
		S2S-comp (\%) & 92.5 & 79.2 & 71.7 & -\\
		\hline
	\end{tabular}
	
	\caption{User evaluation results of our methods against ground truth and other baselines.}
	\label{tab:user_scores}
\end{table}

\textbf{User Study}
To evaluate the perceptual realism and user preference between different methods, we conduct a user study with 53 participants involved in two different experiments: 1) A Just Noticeable Difference (JND) test that ask a user to discriminate a pair of generated/ground truth video clips of varying length between 3$\sim$5 seconds. 2) A side-by-side comparison (S2S-comp) of generated video clips from our INTERnet and a randomly chosen baseline, with source image and driving clip provided for reference. Evaluation scores in Table~\ref{tab:user_scores} indicates superior perceptual realism and overall quality against existing baselines.


\subsection{Ablation Study}
\label{sec:ablation}

We design two sets of ablation tests to verify the efficacy of proposed modules. First, we compare our intrinsic scheme with the other two schemes described in Fig.~\ref{fig:intrinsicreg}. Second, we analyze the importance of temporal regulation for high-resolution case by replacing the multi-scale fusion generator with a single-scaled version \textbf{SS4x}, so that only base motion field at $128\times128$ is required for frame generation. Quantitative scores of five ablation tests are reported in Table~\ref{tab:scores}, where our intrinsic temporal regularization outperforms existing schemes, leading to $2\sim3$ times perceptual gain against the basic SS4x a) setting. Moreover, a significant 0.8 dB gain at frame interpolation is achieved by using intrinsic regulation for multi-scale generator, but there is no gain observed under single-scale setting. This strongly verifies the necessity of our intrinsic scheme for regulating large-scale and high-velocity motion field for high-resolution human video synthesis.

\begin{table*}
	\centering
	\begin{tabular}{|c|ccc|cc|}
		\hline
		& \multicolumn{3}{c|}{\textbf{Per-frame}} & \multicolumn{2}{c|}{\textbf{Temporal}}\\
		Metric & MS-SSIM $\uparrow$ & FID $\downarrow$ & LPIPS $\downarrow$ & Interp-PSNR  $\uparrow$ & Act-Recog-L1  $\downarrow$\\
		\hline
		LWGAN~\cite{liquidwarpinggan}  & 0.677 & 129.152 & 0.189 & 16.977 & 139.061 \\
		GFLA~\cite{GFLA}  & 0.879 & 52.298 & 0.066 & 22.886 & 111.600\\
		FoM~\cite{first-order-motion}  & 0.895 & 91.577 & 0.121 & 23.048 & 97.722 \\
		\hline
		SS4x a) & 0.879 & 87.718 & 0.098 & 23.298 & 97.916 \\
		SS4x b) & 0.911 & 74.399 & 0.079 & 23.056 & 96.645 \\
		SS4x c)  & 0.919 & {73.915} & {0.075} & 23.025 & 91.918 \\
		MS b) & \underline{0.951} & \underline{41.801} & \underline{0.041} & \underline{23.412} & \underline{76.358} \\
		\textbf{Ours (MS c)} & \textbf{0.960} & \textbf{36.451} & \textbf{0.035} & \textbf{24.245} & \textbf{72.844} \\
		\hline
	\end{tabular}
	
	\caption{Quantitative results of different video generation methods on self-imitation task. The best and second best result for each metric are marked bold and underline respectively. Up arrow means a higher score is preferred, and vice versa.}
	\label{tab:scores}
	
\end{table*}

To better illustrate the impact of each individual module, we compare several qualitative examples in Fig.~\ref{fig:ablationv1} where anchor points selected from the base motion estimator are visualized in randomly-colored dots. Clearly, our intrinsic learning scheme helps improve the structural integrity of generated frames, like limbs and body contour, and the reason for such improvement is well justified from the anchor distribution: All anchors predicted with intrinsically-regularized models are close to the motion boundary, whereas two non-intrinsic models produce bad anchor points (green and blue) in the plain background with zero motion. We attribute this to the shortcut created through our intrinsic confidence map, so that temporal loss gradients can be directly back-propagated into the base motion estimator. This could lead to better training dynamics by avoiding possible gradient vanishing/exploding issues, and offering a higher chance of escaping bad local minima. With the stabilized motion field, the multi-scale fusion generator can more accurately transfer low-level features under the driving pose, leading to better preservation of complex textural and geometric patterns (row 1 and 3).

\begin{figure}
	\centering
	\includegraphics[width=1.0\linewidth]{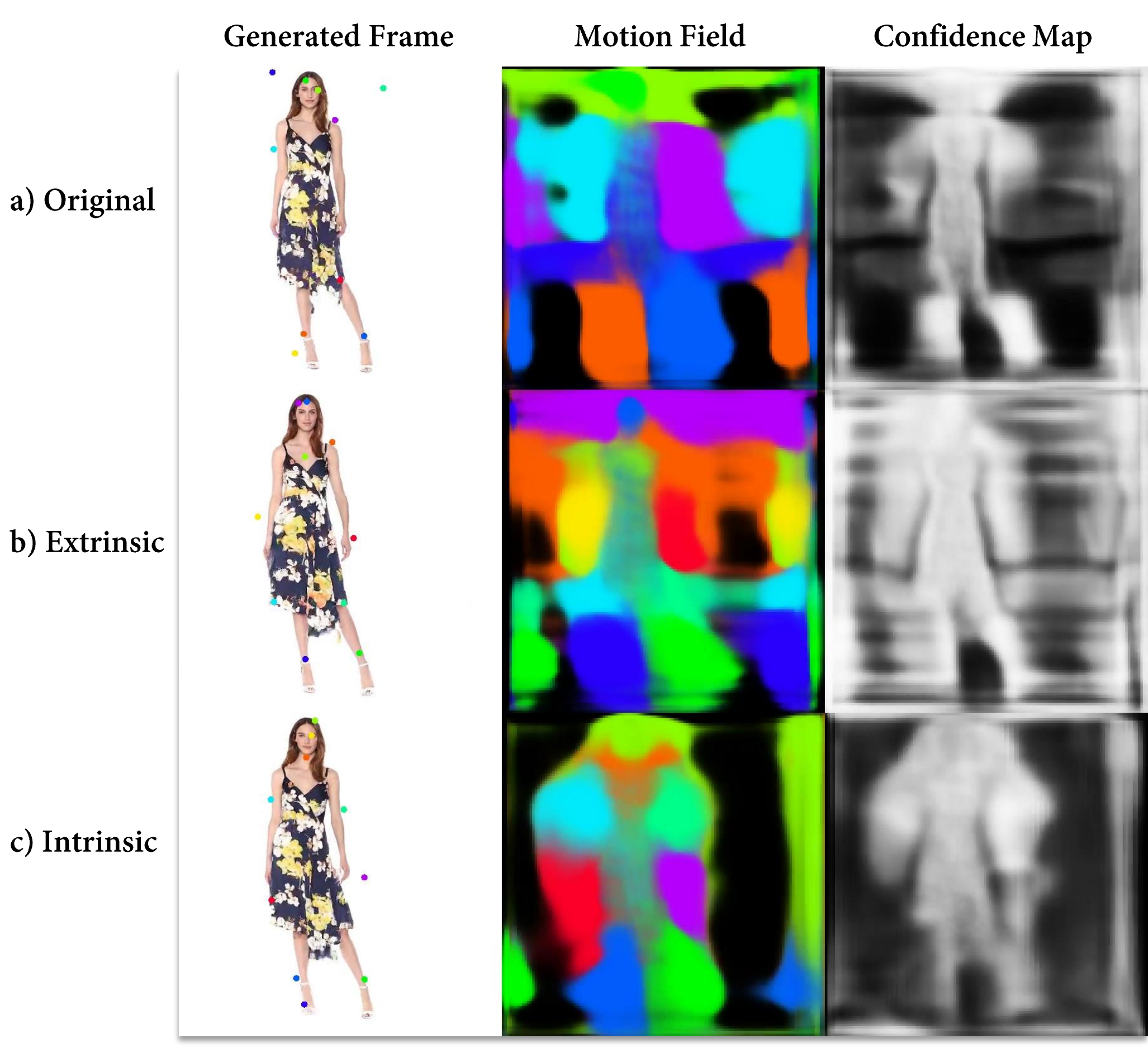}
	\caption[ablation]{Comparison of estimated motion field and occlusion map between different temporal regulation schemes. Each colored blob denotes a local motion region associated to the anchor point with the same color, and the black region indicates still background with no motion.}
	\label{fig:flow_occ}
\end{figure}

\subsection{Visualization of Intrinsic Confidence Map}
To analyze the efficacy of our proposed intrinsic regulation scheme, we visualize the estimated motion field and confidence map of three different regularization schemes, with local motion fields colored according to their anchor points and still background represented as black regions. Here we choose SS4x a), MS b) and MS c) for comparison.\footnote{Note the confidence map is still used for frame generation in a) and b), even if it's not used in temporal loss calculation, see Eqn.~\eqref{eqn:feat_warp}.} As shown in Fig.~\ref{fig:intrinsicreg}, our intrinsic scheme produces more accurate confidence map that better fits the human body structure, and with less background responses.
This helps the motion field estimator to better identify and avoid still backgrounds. Furthermore, it leads to more localized motion field with only a single connected component around anchor points that well correspond to human body parts, while other regulation schemes tend to generate multiple disconnected pieces. Overall, our intrinsic temporal regulation scheme helps with the training of motion field estimator, leading to more stabilized image feature transfer and better temporal smoothness in generated human videos.


\section{Conclusion}
Temporal consistency learning has achieved much progress in boosting single-image processing tasks to the video domain, including style transfer, video-to-video translation, etc. Yet for human image generation, the critical assumption of motion alignment between source and target video is violated, making existing temporal regulation schemes less applicable in this domain. In this paper, we propose an intrinsic temporal regularization scheme to address this issue, which involves a warping loss modulation via a predicted intrinsic confident map. This creates a shortcut for back-propagating temporal loss gradients directly to the motion field estimator, improving its training dynamics and estimation stability.
The proposed regularization scheme can effectively boost existing human image generation frameworks with no extra inference cost, leading to realistic human action videos up to $512\times512$ resolution with better texture preservation and less temporal flickering. In the future, we plan to explore more implementation options of intrinsic temporal consistency learning, such as multi-scale intrinsic warping loss and cross-identity optical flow estimation. Also, we shall verify the efficacy of this work in other non-aligned video generation tasks, \emph{e.g.} face reenactment and dynamic texture rendering (fire, water, cloud, etc.)

{\small
	\bibliographystyle{ieee_fullname}
	\bibliography{cvpr}
}

\end{document}